\documentclass[sigconf]{acmart}

\AtBeginDocument{%
  \providecommand\BibTeX{{%
    \normalfont B\kern-0.5em{\scshape i\kern-0.25em b}\kern-0.8em\TeX}}}

\usepackage{pifont}
\usepackage{microtype}
\usepackage{graphicx}
\usepackage{xspace}
\usepackage{subcaption}
\usepackage{booktabs} 
\usepackage{tabularx}
\usepackage{multirow}
\usepackage{customstyle}
\usepackage{tabulary}
\usepackage{amsmath}
\usepackage{tikz}
\usepackage{xcolor}
\usepackage{pgfplots}
\usepackage{amsthm}
\usepackage{xfrac}

\pgfplotsset{compat=1.14}

\usetikzlibrary{pgfplots.groupplots}
\usepgfplotslibrary{fillbetween}

\usetikzlibrary{positioning, fit, arrows.meta, shapes}

\pgfkeys{%
  /tikz/on layer/.code={
    \pgfonlayer{#1}\begingroup
    \aftergroup\endpgfonlayer
    \aftergroup\endgroup
  }
}
\pgfplotsset{
    highlight/.code args={#1:#2}{
        \fill [every highlight] ({axis cs:#1,0}|-{rel axis cs:0,0}) rectangle ({axis cs:#2,0}|-{rel axis cs:0,1});
    },
    /tikz/every highlight/.style={
        on layer=\pgfkeysvalueof{/pgfplots/highlight layer},
        blue!20
    },
    /tikz/highlight style/.style={
        /tikz/every highlight/.append style=#1
    },
    highlight layer/.initial=axis background
}

\newcommand{\review}{\textcolor[rgb]{0,0,0}}

\newcommand{\sintel}{\textit{Sintel}\xspace}
\newcommand{\system}{\textit{Sintel}\xspace}

\usepackage{colortbl}

\usepackage{colortbl}
\usepackage{xcolor}
\PassOptionsToPackage{svgnames,x11names,dvipsnames}{xcolor}
\usepackage[most]{tcolorbox}

\definecolor{yback}{HTML}{fff2cc}
\definecolor{gback}{HTML}{d9ead3}
\definecolor{bback}{HTML}{c9daf8}
\definecolor{cframe}{HTML}{B9C4CA}
\newtcbox{\inlineboxyellow}[1][]{enhanced,
 box align=base,
 nobeforeafter,
 colback=yback,
 colframe=cframe,
 size=small,
 left=0pt,
 right=0pt,
 boxsep=2pt,
 #1}

\newtcbox{\inlineboxgreen}[1][]{enhanced,
 box align=base,
 nobeforeafter,
 colback=gback,
 colframe=cframe,
 size=small,
 left=0pt,
 right=0pt,
 boxsep=2pt,
 #1}
 
\newtcbox{\inlineboxblue}[1][]{enhanced,
 box align=base,
 nobeforeafter,
 colback=bback,
 colframe=cframe,
 size=small,
 left=0pt,
 right=0pt,
 boxsep=2pt,
 #1}
 
\usepackage{hyperref}

\usepackage{algorithm2e}       

\copyrightyear{2022}
\acmYear{2022}
\setcopyright{rightsretained} 

\acmConference[SIGMOD '22]{Proceedings of the 2022 International Conference on Management of Data}{June 12--17, 2022}{Philadelphia, PA, USA}
\acmBooktitle{Proceedings of the 2022 International Conference on Management of Data (SIGMOD '22), June 12--17, 2022, Philadelphia, PA, USA}
\acmDOI{10.1145/3514221.3517910}
\acmISBN{978-1-4503-9249-5/22/06}

\usepackage{etoolbox}
\makeatletter
\patchcmd{\maketitle}{\@copyrightpermission}{
  \begin{minipage}{0.4\columnwidth}
     \href{http://creativecommons.org/licenses/by/4.0/}{\includegraphics[width=0.95\textwidth]{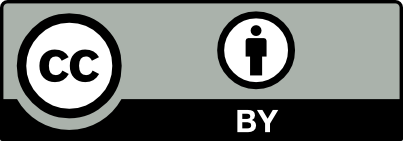}}
  \end{minipage}\hfill
  \begin{minipage}{0.6\columnwidth}
     \href{http://creativecommons.org/licenses/by/4.0/}{This work is licensed under a Creative Commons Attribution International 4.0 License.}
  \end{minipage}
 
  \vspace{5pt}
}{}{}

\makeatother


\begin{document}
\fancyhead{} 

\title{Sintel: A Machine Learning Framework to Extract Insights from Signals}



\author{Sarah Alnegheimish}
\email{smish@mit.edu}
\affiliation{\institution{MIT}%
}

\author{Dongyu Liu}
\email{dongyu@mit.edu}
\affiliation{\institution{MIT}%
}

\author{Carles Sala}
\email{csala@csail.mit.edu}
\affiliation{\institution{MIT}%
}

\author{Laure Berti-Equille}
\email{laure.berti@ird.fr}
\affiliation{\institution{IRD}%
}

\author{Kalyan Veeramachaneni}
\email{kalyanv@mit.edu}
\affiliation{\institution{MIT}%
}

\renewcommand{\shortauthors}{Alnegheimish, et al.}


\begin{abstract}
The detection of anomalies in time series data is a critical task with many monitoring applications. 
Existing systems often fail to encompass an end-to-end detection process, to facilitate comparative analysis of various anomaly detection methods, or to incorporate human knowledge to refine output.
This precludes current methods from being used in real-world settings by practitioners who are not ML experts.
In this paper, we introduce \sintel, a machine learning framework for end-to-end time series tasks such as anomaly detection. The framework uses state-of-the-art approaches to support all steps of the anomaly detection process. \sintel logs the entire anomaly detection journey, providing detailed documentation of anomalies over time. It enables users to analyze signals, compare methods, and investigate anomalies through an interactive visualization tool, where they can annotate, modify, create, and remove events. Using these annotations, the framework leverages human knowledge to improve the anomaly detection pipeline. 
We demonstrate the usability, efficiency, and effectiveness of \sintel through a series of experiments on three public time series datasets, as well as one real-world use case involving spacecraft experts tasked with anomaly analysis tasks.
\sintel's framework, code, and datasets are open-sourced at \url{https://github.com/sintel-dev/}.

\end{abstract}




\maketitle

\section{Introduction}



Many industries have set up processes and workflows for analyzing time series to monitor, control, and optimize the behaviours of the products and services they offer. These range from monitoring applications that we use on a daily basis (Zoom, Slack, and several others) to managing a fleet of heavy equipment like turbines and satellites. These workflows usually rely on data visualization, domain knowledge and some simple rule-based decision making. Even with the proliferation of Machine Learning (ML) in vision and language systems, integration of ML into these workflows is still an open problem. In this paper, we pick one amongst those workflows -- time series anomaly detection -- and present the challenges and solutions. 

Effective Anomaly Detection (AD) methods can identify deviations from normal behavior and notify users, sounding the alarm about potential problems.
Researchers have been developing these methods for decades~\cite{Gupta2014}. 
As data has become larger, more complex, and increasingly multidimensional, traditional distance- \cite{Chaovalitwongse2007}, density- \cite{Zhang2007} or isolation-based \cite{isolationforest2008} approaches have begun to perform less competitively in real-world scenarios.

Since then, ML and deep learning based methods have garnered increased attention~\cite{AAAI19,KDD19,ICDM18,jacob2021exathlon}.
With the wide diversity of available methods, choosing pipelines and evaluating which one is most suitable can be difficult.
Depending on the application, it may be insufficient to rely on data benchmarks~\cite{Numenta15} to determine the best methods, particularly when there is no ground truth. Ideally, it would be possible to test and compare the performances of all methods on a data of interest. 
Services and libraries, such as Microsoft  Azure~\cite{microsoft2019time}, and Luminaire~\cite{chakraborty2020luminaire}, for AD have been recently proposed. 
However, their users may still run into the following problems.

\textbf{HIL analysis}. Services often neglect the Human-In-the-Loop (HIL) dimension, which is vital both for validating detected anomalies and for differentiating between true errors and legitimate exceptions.
To confirm and compare anomalies, users --- generally domain experts, machine learning researchers, or data scientists --- resort to visualisation to inspect signals and view different aggregation levels.
Usually, users need to shift to a programming language that they are comfortable with (e.g., \texttt{MATLAB}, \texttt{R}, or \texttt{Python}) to complete this process, which could result in loss of information.

\textbf{Contextual knowledge}. Many existing solutions consider individual time series in isolation -- even though in most real-world situations, thousands of multivariate time series are correlated and monitored continuously. For instance, detecting collective and correlated anomalies across complex time series is often important in health monitoring.
By bringing all the information into a single platform, we create a thorough knowledge base.

\textbf{Customizing workflows}. Users may want to customize or compose an anomaly detection system for their own use cases. However, designing a platform to analyze a specific type of time series, detect anomalies therein, and integrate domain knowledge into the resulting validation is complicated, and there is no solution that supports this type of technical workflow. Such a solution would require a systematic definition of procedures and tasks, a comprehensive set of application programming interfaces (APIs), a modular and extensible machine learning pipeline design, and an interactive investigation of anomalies.


\definecolor{mygray}{gray}{0.8}
\newcommand{\cmark}{\ding{51}}%
\newcommand{\xmark}{\color{mygray}\ding{55}}%

\newcolumntype{M}[1]{>{\centering\arraybackslash}m{#1}}
\begin{table*}[!t]
\begin{tabular}{l|l|M{4em}*{2}{M{3.5em}}M{3em}M{4em}M{3em}*{3}{M{3.5em}}|c}
\toprule
\multicolumn{2}{c|}{}                        
& MS Azure \cite{microsoft2019time}
& ADTK \cite{adtk}
& Luminaire \cite{chakraborty2020luminaire}
& TODS \cite{lai2020tods}
& Telemanom \cite{Hundman2018}
& NAB \cite{ahmad2017unsupervised}
& EGADS \cite{Laptev2015}
& Stumpy \cite{law2019stumpy}
& GluonTS \cite{gluonts_jmlr}
& \sintel \\

\midrule
\multirow{3}{*}{\rotatebox[origin=c]{90}{Users}} 
& End User 
& \cmark    
& \cmark 
& \cmark
& \xmark
& \xmark                            
& \xmark
& \xmark
& \cmark   
& \xmark 
& \cmark   \\
& System Builder 
& \cmark    
& \xmark
& \xmark
& \xmark 
& \xmark                           
& \xmark
& \xmark
& \xmark
& \xmark
& \cmark   \\
& ML Researcher
& \xmark
& \xmark
& \xmark
& \cmark   
& \cmark                             
& \cmark
& \cmark
& \xmark
& \cmark   
& \cmark   \\
\midrule
\multirow{3}{*}{\rotatebox[origin=c]{90}{Engine}}    
& Preprocessing
& \xmark
& \cmark
& \cmark
& \cmark
& \xmark                        
& \xmark
& \xmark
& \cmark   
& \cmark   
& \cmark   \\
& Modeling
& \cmark    
& \cmark 
& \cmark
& \cmark   
& \cmark                             
& \cmark
& \cmark
& \xmark
& \cmark   
& \cmark   \\ 
& Postprocessing 
& \xmark
& \cmark
& \cmark
& \cmark
& \xmark
& \xmark
& \xmark
& \cmark   
& \xmark
& \cmark   \\
\midrule
\multicolumn{2}{l|}{Modular}
& \xmark
& \cmark
& \cmark
& \cmark
& \xmark
& \xmark
& \xmark
& \cmark   
& \cmark   
& \cmark   \\
\midrule
\multirow{3}{*}{\rotatebox[origin=c]{90}{Comp.}}
& Evaluation
& \xmark
& \cmark
& \xmark
& \xmark
& \cmark                             
& \xmark
& \xmark
& \xmark
& \xmark
& \cmark   \\
& Benchmark
& \xmark
& \xmark
& \xmark
& \cmark   
& \xmark
& \cmark
& \xmark
& \xmark   
& \cmark   
& \cmark   \\
& Database
& \cmark    
& \xmark
& \xmark
& \xmark   
& \xmark
& \xmark
& \xmark   
& \xmark   
& \xmark   
& \cmark   \\
\midrule
\multirow{2}{*}{\rotatebox[origin=c]{90}{API}} 
& \texttt{lang. specific}
& \cmark    
& \cmark
& \cmark
& \cmark   
& \xmark                            
& \cmark
& \xmark
& \cmark   
& \cmark   
& \cmark   \\
& \texttt{RESTful}
& \cmark    
& \xmark
& \xmark
& \xmark   
& \xmark                            
& \xmark
& \xmark
& \xmark   
& \xmark   
& \cmark   \\
\midrule
\multicolumn{2}{l|}{HIL}
& \xmark    
& \xmark
& \xmark
& \xmark   
& \xmark                            
& \xmark
& \xmark
& \xmark   
& \xmark   
& \cmark   \\
\bottomrule
\end{tabular}
\caption{Comparison of anomaly detection software. A (\ding{51}) indicates the package includes an attribute, while an ({\color{mygray}\ding{55}}) indicates the attribute is absent. Attribute categories from top to bottom: \textit{Users} shows which user types can benefit from each software: \textit{end users} are interested in detecting anomalies, \textit{system builders} are interested in adding their own workflows, and \textit{ML researchers} are interested in creating new pipelines that outperform existing methods; \textit{Engine} denotes the operations handled by each software; \textit{Modular} indicates whether or not pipelines can reuse primitives; \textit{Comp.} shows whether systems include certain components including custom \textit{evaluation} mechanisms, \textit{benchmarking} frameworks, and an integrated \textit{databases} of results; \textit{API} refers to the inclusion of APIs for user interaction, whether language-specific or \texttt{REST}; and \textit{HIL} denotes the presence or absence of a human-in-the-loop component that can integrate experts' knowledge back into the system.}
\label{tab:related_work}
\vspace{-0.7cm}
\end{table*}

To address these problems,
we introduce \system.
The framework tackles the problem of anomaly detection end-to-end, from the first step of time series ingestion through machine learning modeling, interactive visualization, and user feedback. It is a comprehensive, streamlined ecosystem that targets various user needs.

The key contributions of this paper are summarized as follows:
\begin{itemize}
    \item \textbf{An efficient end-to-end time series anomaly detection framework.} \sintel provides a suite of anomaly detection pipelines executable through a user-friendly interface. Users kickstart the system by presenting a signal, which trains a model and returns the detected anomalies. The framework's modular nature facilitates the creation, exchange and reuse of primitives between different pipelines. 
    \item \textbf{An ecosystem for anomaly interaction and annotation-based learning.} The framework includes a human-in-the-loop component, allowing domain experts to properly annotate and interact with detected anomalies. We support this with a visualization tool that aids users in the inspection and investigation processes. The feedback component within the framework learns from human annotations to further improve the detection performance. 
    \item \textbf{A standardized benchmarking framework for time series anomaly detection pipelines.} We designed a comprehensive benchmarking suite to compare multiple pipelines on a collection of different time series datasets. The benchmarking suite features intricate evaluation metrics designed specifically for anomaly detection. This feature enables users to run multiple experiments under the same conditions in order to obtain fair, empirical comparisons. As of writing the paper, the benchmarking suite has 6 pipelines (1 statistical and 5 deep learning models) and 2 evaluation mechanisms.
    \item \textbf{A comprehensive evaluation.} We evaluate our framework by benchmarking all pipelines on a collection of 11 datasets from three reputable data sources -- NASA, Yahoo, and Numenta -- and reporting their quality and computational performance. Moreover, we conduct an experiment to assess the framework's ability to learn from experts' annotations. We compare the features of \sintel against existing systems (presented in Table~\ref{tab:related_work}). Lastly, we deploy the \system in a real-world setting with a leading satellite operation company.
    \item \textbf{Open-source.} Our framework, code and datasets are available at: \href{https://github.com/sintel-dev/Orion}{Orion}, \href{https://github.com/sintel-dev/Draco}{Draco}, \href{https://github.com/sintel-dev/MTV}{MTV}, and \href{https://github.com/sintel-dev/sintel}{sintel API}. Scripts to reproduce the results in the paper are available in~\cite{sintel_paper_smish}.
\end{itemize}

\section{Sintel}
Throughout this section, we instantiate the core idea for time series anomaly detection.
Given $X= \{x_1, x_2, x_3, \dots, x_n\}$, a multivariate signal with $m$ channels where $x_i \in \mathbb{R}^m$, and assuming there exists a set of \textit{variable-length} anomalies $A = \{(t_s, t_e)\:|\: 1 \leq t_s < t_e \leq n\}$ that is \underline{unknown} a priori, \sintel aims to detect $A$ using a combination of machine and human intelligence.
Note that \sintel addresses \textit{variable-length} anomalies. For two anomalous intervals $a_1 = (t^1_s, t^1_e)$ and $a_2 = (t^2_s, t^2_e)$, the length of $a_1$ 
does not necessarily equal the length of $a_2$.
The system comprises a series of components that can perform all the necessary detection steps, from composing pipelines to annotating anomalies. We describe the details in this section.

\vspace{-0.4cm}
\subsection{Real-World Scenario}\label{sec:example_scenario}
\begin{figure}[!t]
    \centering
    \includegraphics[width=1.0\linewidth]{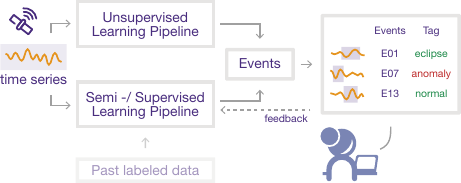}
    \caption{AD workflow. Use an unsupervised pipeline to locate anomalies, which are then presented to the expert for annotation. The annotated events are provided to the semi-/supervised pipeline --- which can be pre-trained with past labeled data --- to learn from feedback and keep improving.}
    \label{fig:ad_workflow}
    \vspace{-0.6cm}
\end{figure}

To explain the motivation for \system and illustrate the anomaly detection workflow (Figure~\ref{fig:ad_workflow}), we describe a real-world scenario drawn from our three-year collaboration with a world-leading communication satellite company. One major objective of the company's operative team is to detect unexpected behaviors (i.e., anomalies) in tens of thousands of signals. We collaborated with a spacecraft program manager and 5 senior satellite engineers, each of whom has considerable experience in telemetry data analysis (between 5 and 17 years), but relatively little ML experience (0 to 3 years). 

The team works with multiple spacecrafts. Each spacecraft telemetry database contains around 37K signals spanning 9 different subsystems. Each signal is a univariate time series collected at microsecond level, and has been tracked for over 10 years.
The team's conventional approach to anomaly detection is based around setting and adjusting thresholds in order to flag anomalous intervals.
The team then reviews the suspicious intervals manually, often using simple \texttt{csv} files, and examines individual signals in a third-party platform such as \texttt{MATLAB}.
Around 20 alarms are reported every day, most of which can be resolved within a few hours. 
For some that are identified as true alarms but non-urgent, the experts gather further information over some time window to help explain the root cause and the way forward.

We experienced similar challenges with other teams as well.
Over the course of this process, teams face the following challenges: 
(\textbf{C1}) Setting and adjusting threshold methods can be user demanding and laborious, forcing teams to restrict their focus to a subset of a few hundred signals chosen based on domain knowledge;
(\textbf{C2}) Teams want to use ML models to identify contextual anomalies --- anomalies that do not exceed a normal range, but are unusual compared to local values. With limited ML experience, it can be difficult to adopt such methods;
(\textbf{C3}) With the abundance of AD methods, teams struggle to know which ML model to select for a particular dataset;
(\textbf{C4}) Teams found that ML models often flag unusual patterns, even when these patterns do not necessarily indicate a problem. For example, a maneuver might cause patterns that are then flagged even though they are not troublesome. Teams are eager for their models to ignore these patterns, but struggle to integrate this feedback;
(\textbf{C5}) Teams frequently discuss anomalous events of interest. However, they are not currently collecting these events into a knowledge base to learn from.

\sintel addresses all of these challenges.
\sintel is an automated end-to-end framework that is able to identify anomalous events from tens of thousands of signals (\textbf{C1}).
It integrates various state-of-the-art anomaly detection pipelines, and provides simple user-friendly APIs to interact with the framework (\textbf{C2}).
With the benchmark suite, \sintel provides results in an organized manner.
This helps the team to easily pick a suitable, existing, and verified unsupervised pipeline from our collection, or even to create their own (\textbf{C3}).
Through \sintel's visual interface, the team investigates flagged anomalies, annotates them, and incorporates the feedback back into the framework (\textbf{C4}). 
The feedback is stored in an integrated database that maintains the results of each detected event and annotations (\textbf{C5}). 

\subsection{Anomaly Detection Pipelines}
\label{sec:anomaly_pipeline}
Anomaly detection pipelines take a univariate or a multivariate signal as input and use it to generate an array of intervals $A = [(t^1_s, t^1_e), \cdots, (t^k_s, t^k_e)]$ representing the anomalies discovered. In most cases, users lack labels for their data. Therefore, our pipelines service both unsupervised and supervised approaches, and are built end-to-end such that \sintel is agnostic as to which pipeline is executed.
To understand the composition of pipelines, we describe their basic building blocks, or \textit{primitives}, below.

\textbf{Primitives} are reusable software components~\cite{smith2019machine}.
A primitive receives data in the form of a specified input, performs an operation, and returns a calculated output.
Each primitive is responsible for a single task, ranging from data transformation to signal processing to machine learning modeling to error calculation.
It is possible to build complex pipelines by stacking primitives on top of one another.
Each primitive has associated metadata including annotations, such as the name of the
primitive, the description and documentation link, and the engine category.
As illustrated in Table~\ref{tab:related_work}, \system covers three \review{engines}:

\noindent\inlineboxyellow{\textbf{Pre-processing}}
Time series are rarely handled in their raw form. Before using a signal, the data must be transformed through pre-processing, such as imputing missing values. 

\noindent\inlineboxgreen{\textbf{Modeling}}
Once the signal has been processed, we can start modeling it.
There are different techniques for modeling. In time series anomaly detection, we are interested in predicting or reconstructing the signal so that we can have an \textit{expected} signal.

\noindent\inlineboxblue{\textbf{Post-processing}}
After generating an \textit{expected} signal, we use discrepancies between the expected and the real signal to find anomalies. We refer to this process as error calculation. Post-processing primitives output intervals containing potentially anomalous sub-sequences alongside their likelihood probability of being anomalous.

Modularly designed engines can re-use primitives between, within and across pipelines.
This reduces the number of lines of code -- and thus error potential -- and increases transparency.
Having a granular definition also encourages best practices such as proper documentation, unit tests, and validation.
Contributors can integrate a new primitive into \system without modifying an entire pipeline.

\begin{figure}[!t]
\includegraphics[width=0.81\linewidth]{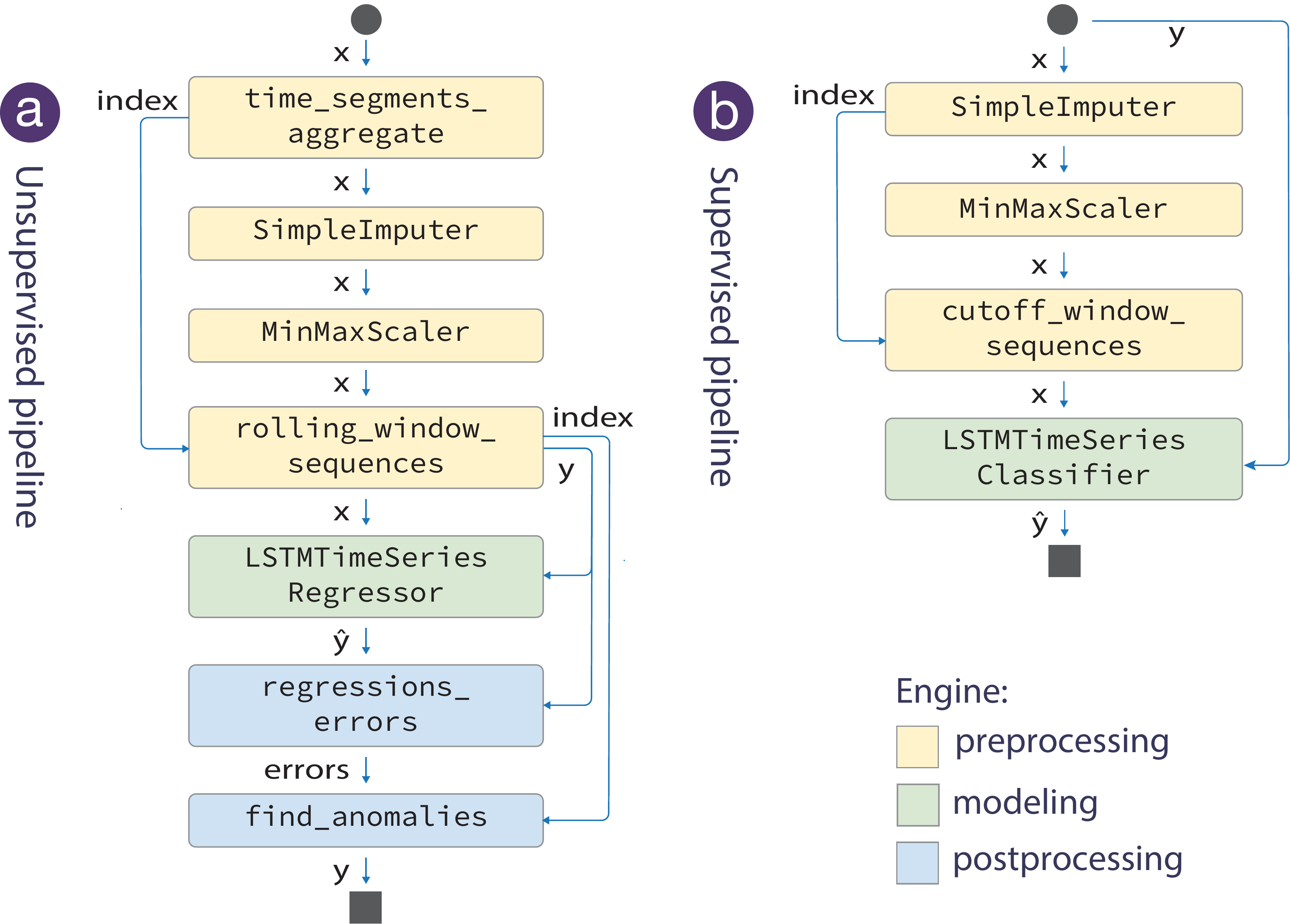}
\vspace{-0.3cm}
\caption{
Graphic representations of
(a)  \review{an unsupervised} pipeline with LSTM network.
(b) \review{its supervised counterpart}.
}
\label{fig:orion-api}
\vspace{-0.5cm}
\end{figure}

\textbf{Pipelines}\label{sec:pipelines}
are end-to-end programs composed of primitives. 
Each pipeline is computed into its respective computational graph similar to the examples shown in Figure~\ref{fig:orion-api}.
In this paper, the term ``pipeline'' always refers to a program tasked to identify anomalies in time series data.
Primitive and pipeline structures have been successfully adopted in many applications, including healthcare~\cite{alnegheimish2020cardea, smith2019machine}.

\paragraph{Dissecting LSTM Pipeline}
The pipeline in Figure~\ref{fig:orion-api}a uses a Long Short-Term Memory (LSTM) network to predict data values at future timestamps. 
We first take a raw signal $\mathbf{x}$ and feed it into the \texttt{time\_segments\_aggregate} to produce $\mathbf{x} = [x^1, x^2, \dots, x^T]$ where the time intervals between $x^{t-1}$ and $x^t$ are equal. 
Then we scale the data $\mathbf{x} \in [-1, 1]$ and impute missing values using the mean value of the signal.
After that, we create the training window sequences.
The processed signal is now ready to train the double-stacked LSTM network.
Once the network is trained, we generate the predicted signal and compute the discrepancies using \texttt{regression\_errors}, which is an absolute point-wise difference $| \mathbf{\hat{x}} - \mathbf{x} |$.
Lastly, we use dynamic threshold on error values to find anomalous regions~\cite{Hundman2018}.
Customizing pipelines is fairly easy. Users can configure a primitive or even replace it with another. For example, to use z-score normalization, users can swap \texttt{SimpleImputer} with \texttt{StandardScaler}.

\begin{algorithm}
\SetKw{KwReturn}{return}
\KwIn{ground truth anomalies $T$, predicted anomalies $P$}
\KwOut{confusion matrix}
 \SetKwBlock{Begin}{begin}{}
 \SetKwFunction{KwFunction}{confusion\_matrix}
 \Begin{
  $E \leftarrow T \cup P$ \hfill \tcp{all $t_s$ and $t_e$ timestamps}
  $\tilde{T} \leftarrow \varnothing, \tilde{P} \leftarrow \varnothing, W \leftarrow \varnothing$\;
  $E \leftarrow$ sort($E$) \hfill \tcp{sort timestamps from small to large}
  $s \leftarrow$ pop($E$) \hfill \tcp{the first timestamp}
  \While{$E \neq \varnothing$}{
   $e \leftarrow$ pop($E$)\;
   $ti \leftarrow \review{(t_s, t_e)}$ \hfill \tcp{create a time interval \review{($t_s$, $t_e$)}}
   $W \leftarrow W \cup \review{\{t_e - t_s\}}$\;
   $\tilde{T} \leftarrow \tilde{T} \cup \{\text{overlap}(ti, T)\}$ \hfill \tcp{check if $ti$ in ground truth}
   $\tilde{P} \leftarrow \tilde{P} \cup \{\text{overlap}(ti, P)\}$ \hfill \tcp{check if $ti$ in predicted} 
   $s \leftarrow e$\;
  }
 \KwReturn{{\normalfont $\KwFunction\:(\tilde{T}, \tilde{P}, W)$}}
 }
 \caption{Weighted Segment Evaluation. We create sequences partitioned based on the ground truth and predicted anomalies. For each sequence, we obtain a time range and whether it is part of ground truth or predicted set. We compute confusion matrix weighted by its respective duration.\vspace{-0.35cm}}
 \label{algo:weighted_segment}
\end{algorithm}


\begin{algorithm}[!t]
\SetKw{KwReturn}{return}
\KwIn{ground truth anomalies $T$, predicted anomalies $P$}
\KwOut{confusion matrix $\langle\text{tp},\text{fp},\text{fn} \rangle$}
 \SetKwBlock{Begin}{begin}{}
 \Begin{
  $U \leftarrow \varnothing$ \hfill \tcp{bookkeeping unmatched events}
  $\text{tp} \leftarrow 0$\;
  \While{$T \neq \varnothing$}{
  $t \leftarrow$ pop($T$)\;
  \For{$p \in P$}{
   \If{{\normalfont overlap($t$, $p$)}}{
    $\text{tp} \leftarrow \text{tp} + 1$ \hfill \tcp{matched}
   }
  }
  \If{{\normalfont unmatched($t$)}}{
   $U \leftarrow U \cup \{t\}$ \hfill \tcp{add to unmatched}
  }
 }
 $\text{tn} \leftarrow |U|$\;
 $\text{fp} \leftarrow |P| - \text{tp}$\;
 \KwReturn{{\normalfont $\langle\text{tp},\text{fp},\text{fn}\rangle$}}
 }
 \caption{Overlapping Segment Evaluation. For each ground truth anomaly, we search whether it overlaps with any event in the predicted set. If so, it counts towards a true positive; if not, it is considered a true negative. Then the total false positives is the complement of true positives.\vspace{-0.51cm}}
 \label{algo:overlapping_segment}
\end{algorithm}

\subsection{Pipeline Evaluation Metrics}
\label{sec:scoring}
In order to measure the relative performances of different pipelines, we need a way to compare detected anomalies with ground truth labels according to a well-defined metric. In classification, the most widely-used metrics include \textit{precision}, \textit{recall} and \textit{F1-score}. However, as noted by \citet{Tatbul2018}, these scores are not useful in the context of time series where data is not regularly sampled. For a given set of ground truth anomalies $T = \{(t_s, t_e)\}^{m}_{i=1}$ and predicted anomalies $P = \{(t_s, t_e)\}^{n}_{i=1}$ where $t_s$ and $t_e$ represent the start and end timestamps of an anomaly respectively, we define specific methods to enable the fair computation of metrics without restrictions on the data: weighted segment and overlapping segment. Each approach allows for a different assessment.

\subsubsection{Weighted Segment}
Weighted segment-based evaluation is a strict approach that weights each segment according to its actual time duration. As illustrated in Algorithm~\ref{algo:weighted_segment}, the time series is segmented into multiple sequences by the edges of the anomalous intervals. For each segment, we compute its confusion matrix and weight it by its time range. This approach is valuable when the user aims to detect the exact segment of the anomaly. In cases where the user's signal is inherently regularly sampled, this approach is equivalent to a sample-based evaluation. 

\subsubsection{Overlapping Segment}
Overlapping segment is a more lenient evaluation approach. It is inspired by the evaluation method of \citet{Hundman2018}, which rewards the model if it alerts the user to even a subset of an anomaly. This is considered sufficient because domain experts monitor the signal and will investigate even an imprecise alarm, likely discovering the full anomaly. Algorithm~\ref{algo:overlapping_segment} illustrates our approach to counting true positives, false positives, and false negatives.

\subsection{Human Annotations}
Once the first phase of anomaly detection using machine learning is complete, we require human expertise to validate and annotate the identified anomalies, which may or may not be based on ground truth. We provide a visualizing subsystem so that experts can view time series and their respective predicted anomalies (see Section~\ref{sec:viz}).
Users can interact with this system by \textit{confirming, modifying, removing, searching,} and \textit{discussing} events.

\textit{Why is the logic important?}
These annotations effectively incorporate domain-specific knowledge into our anomaly detection framework. This reduces both false positives and false negatives and helps improve future predictions.

\section{System Design and Architecture}

\begin{figure}[!t]
\includegraphics[width=1.0\linewidth]{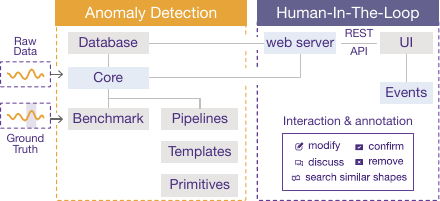}
\vspace{-0.5cm}
\caption{\review{\sintel consists of two subsystems. The anomaly detection subsystem detects anomalies, which are then annotated by users using the human-in-the-loop subsystem.}} 
\label{fig:system}
\vspace{-0.45cm}
\end{figure}

\sintel's components are categorized into two main subsystems: anomaly detection and human-in-the-loop, with a communication channel between them (Figure~\ref{fig:system}). 
The core (Section \ref{sec:core}) is the main entry point, which allows users to train and benchmark pipelines and to predict and store anomalies. The machine learning stack (Section \ref{sec:mlstack}) contains primitives, templates, and pipelines, followed by the description of the hyperparameter tuning component (Section 3.3). Due to the nature of anomaly detection pipelines we include a benchmark utility (Section \ref{sec:cs_benchmarking}) to compare the quality and computational performance of different pipelines. We supplement the framework with a database (Section \ref{sec:db}) to keep a persistent state of information. To incorporate human knowledge, we use the visualization subsystem to allow experts to annotate events and utilize human annotations through the feedback loop (Section \ref{sec:feedback}).

\begin{figure}[!ht]
\includegraphics[width=\linewidth]{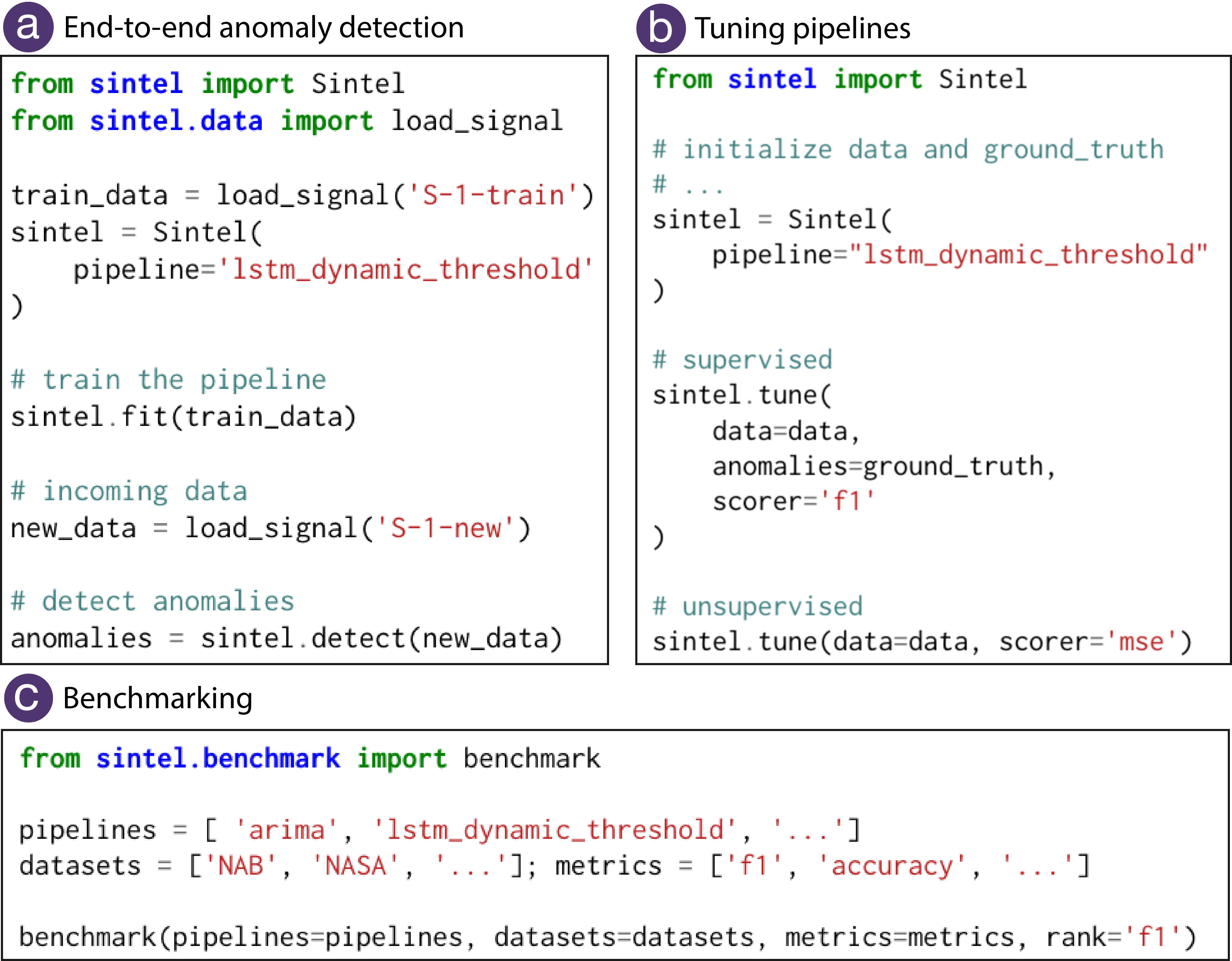}
\vspace{-0.7cm}
\caption{
\review{Example code for (a) end-to-end anomaly detection, where users load the data, select a pipeline (e.g. \texttt{lstm\_dynamic\_threshold}), train a pipeline, and detect anomalies; (b) pipeline tuning; and (c) benchmarking.}
}
\label{fig:orion-code}
\vspace{-0.5cm}
\end{figure}


\vspace{-0.2cm}
\subsection{Core}\label{sec:core}
\sintel's core provides a set of coherent APIs, allowing users to execute end-to-end processes (\textbf{C1, C2}).
Given a signal $X$, we want to obtain a set of detected anomalies. 
With \sintel, this is straightforward.
First, the user loads a signal which follows the input standard -- (timestamp, values).
Next, the user selects the pipeline of interest from a suite of available options. 
Once a pipeline is selected, it is trained on the signal using \texttt{sintel.fit(data)}. 
To detect anomalies, the user then executes \texttt{sintel.detect(data)} to produces a set of possible anomalies.
\review{Figure~\ref{fig:orion-code}a} shows this process.

Simple code execution, accomplished via \texttt{fit/detect/evaluate} functionalities and the pipeline, makes the framework unified, usable, and accessible -- similar to the popular \texttt{fit/predict} interfaces for democratized libraries such as scikit-learn~\cite{buitinck2013scikitlearnapi}.

\subsection{Machine Learning Stack}\label{sec:mlstack}
As described in Section~\ref{sec:pipelines}, a pipeline is a set of primitives that executes an operation. Primitives can be reused between pipelines, which makes \sintel code-efficient. 
In most cases, pipelines require primitive hyperparameters to be set based on the dataset. To satisfy this requirement, we introduce a \textit{template} concept where a template $T = \langle V, E, \Lambda \rangle$, $V$ is a set of pipeline steps, $E$ is a set of edges between steps to represent data flow, and $\Lambda$ is the joint hyperparameter space for the underlying primitives~\cite{smith2019machine}. Following this definition, a pipeline is a configured template with a fixed hyperparameter setting $P = \langle V, E, \lambda \rangle$ where $\lambda \in \Lambda$ is a specific set of hyperparameters. This definition allows us to create and manipulate pipelines easily, enabling their use with a wide range of signals. More importantly, it gives us visibility into which hyperparameters are altered when the pipeline is run on one dataset versus another. 
This transparency is crucial to making our results reproducible.

\paragraph{AD Pipeline Hub}
\sintel stores a collection of end-to-end anomaly detection pipelines that work with state-of-the-art methods.
As of the time of writing, we have incorporated \texttt{LSTM DT}~\cite{Hundman2018}, \texttt{LSTM AE}~\cite{Malhotra2016}, and \texttt{TadGAN}~\cite{geiger2020tadgan}. Moreover, we provide ``pipelines'' that can connect to existing anomaly detection services, such as MS Azure~\cite{microsoft2019time}. This set of pipelines is easily extendable and can be expanded further (\textbf{C2}).

\begin{figure}[!htb]
    \centering
    \includegraphics[width=1.0\linewidth]{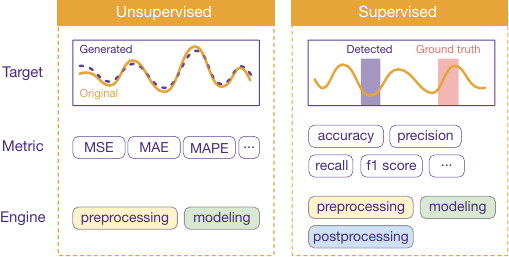}
    \vspace{-0.6cm}
    \caption{\review{Hyperparameter tuning in two conditions: (1) unsupervised, where the goal is to optimize the signal generated by the ML model, and (2) supervised, where our goal is to produce anomalies that best match the ground truth set.}}
    \label{fig:hpo}
    \vspace{-0.5cm}
\end{figure}

\subsection{Tuning Hyperparameters}
\label{sec:hpo}
Hyperparameter tuning is instrumental to ML systems~\cite{Bergstra2012, bergstra2011algorithms}.
To tune pipelines automatically (Figure~\ref{fig:orion-code}b), we integrate \texttt{BTB},
an open-source and extensible framework with black box Bayesian Optimization~\cite{smith2019machine}.
In short, the AutoML component of the framework aims to find the configuration of hyperparameters for a given pipeline template that best maximizes some set of objective functions. Given pipeline template $T$ and an objective function $f$ that assigns a performance score to pipeline $P_\lambda$ with hyperparameters $\lambda \in \Lambda$, we recover $\lambda^\ast = arg max_{\lambda \in \Lambda} f(P_\lambda)$. 
We use \texttt{GPTuner}, which optimizes candidates using a Gaussian process meta-model, records evaluations, and proposes hyperparameters $\lambda$. We continue the search until our budget runs out, or we have reached the optimal value.

\sintel's tuners are customized for use in two different settings, as shown in Figure~\ref{fig:hpo}. In an \textit{unsupervised} setting, a user only tunes the sub-pipeline that attempts to generate the signal closest to the original signal. To achieve this, users specify the pipeline template and evaluation metrics, such as \texttt{MAPE}, \texttt{MAE}, etc. for their objective function.
In a \textit{supervised} setting, we provide objective functions that evaluate the efficacy of the pipeline in detecting known anomalies, such as \texttt{F1}. Note that in the supervised setting, a ground truth set of anomalies must be defined.
Tuning can be used to fine-tune a pipeline with expert annotations (\textbf{C4}). 
Depending on the tuning setting, some engines may not need to be tuned. Recall in Section~\ref{sec:anomaly_pipeline}, our primitives are annotated, enabling \sintel to automatically pull hyperparameters with respect to the set of primitives needed.

\vspace{0.2cm}
\subsection{Benchmarking Framework}\label{sec:cs_benchmarking}
The availability of benchmarking is one of the key advantages of our framework (\textbf{C3}). \review{As shown in Table~\ref{tab:related_work}, many existing frameworks develop algorithms for AD, but lack an out-of-the-box benchmark.} \review{With \sintel's evaluation metrics and pipeline hub}, we are able to thoroughly compare these methods on \review{multiple} datasets, \review{through a single command --- \texttt{sintel.benchmark}. Figure~\ref{fig:orion-code}c illustrates our benchmark API}. The benchmark is designed to measure two main aspects: quality and computational performance.

\subsubsection{Quality performance}
We define quality evaluation as an assessment of how well the pipeline has detected ground truth anomalies. We extend beyond sample-based evaluation metrics such as precision and recall as defined in \review{\href{https://scikit-learn.org/stable/modules/classes.html\#module-sklearn.metrics}{\texttt{scikit-learn}}}, and introduce our pipeline evaluation metrics as detailed in Section~\ref{sec:scoring}. The benchmark is extensible, and users can easily add new metrics and evaluation criteria.

\subsubsection{Computational performance}
Deep learning is notorious for its expensive computational needs. In addition to quality performance, it is important to compare different pipelines' computational costs. To accomplish this, we measure the total time necessary for each pipeline's execution, including how much time is required to train the pipeline (\textit{training time}) and how much time the pipeline takes to turn an input into an output (\textit{pipeline latency}). Additionally, we record how much time each individual primitive needs for both phases. We also measure memory consumption for pipelines and primitives.
The inclusion of computational monitoring allows us to identify which primitives cause bottlenecks within the pipeline and to improve them further.


\subsection{Persistent Knowledge Base}\label{sec:db}
\begin{figure}[!t]
\includegraphics[width=0.6\linewidth]{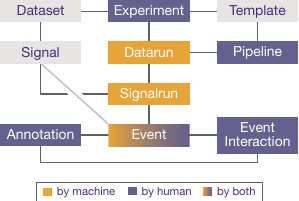}
\vspace{-0.3cm}
\caption{High-level database schema.}
\label{fig:db-schema}
\vspace{-0.65cm}
\end{figure}

In real-world settings, it is often necessary to continuously record data, including expected anomalies (\textbf{C5}). It is also important to document events so that information is not lost, which can lead to repetitive and unnecessary investigations. Proper logging of signals and their corresponding anomalies will greatly improve users' understanding of where these anomalies came from, and tracing back decisions will become easier as well. 
In \system, we use a \texttt{mongoDB} \href{https://github.com/sintel-dev/Orion/blob/master/DATABASE.md}{database} with an extensive schema to fill this gap.
\review{We chose \texttt{mongoDB} due to: (1) its flexibility over application domains; (2) its interoperability with \sintel's visualization tool --- a web app developed using JavaScript; (3) its extensiblility through community add-ons.}
Incorporation of the database enables users to do the following:
(1) use anomalies that exist within the database to annotate new signals, in order to avoid rerunning the same model and wasting computational time;
(2) document the history of anomalies as users become more experienced;
and (3) maintain a growing store of information as multiple users annotate signals.
A high-level depiction of the entities and relationships within the database is shown in Figure~\ref{fig:db-schema}.
Through the database, we can retrieve information easily, constantly trace what is happening, and create a valuable knowledge base for both new and experienced users.

\vspace{-0.2cm}
\subsection{Anomaly Annotation and Feedback}\label{sec:viz}\label{sec:feedback}
Another phase of our framework's workflow is anomaly analysis.
Our goal is to enable the user to annotate flagged events and inspect detected anomalies.
To achieve this, we introduce an interactive visualization tool~\cite{liu2021mtv}.
This tool allows experts to undertake an investigation process, and to annotate and discuss events.


The visualization subsystem supports standard operations such as multi-signal viewing.
In addition, it allows for a multi-aggregation view, which enables a user to compare the signal at different aggregation levels.
These operations help experts understand why certain intervals have been flagged, and allows for modification and annotation.
In addition, we provide a discussion panel so that team members can comment on or dispute the status of an event.

Expert annotations are extremely important for understanding whether certain events are truly anomalous. 
Moreover, these annotations are persistent, allowing future teams and users to understand why certain decisions were made.
Although the sequence of discussions and actions that have led to the classification of an event can help to form the canonical logic behind a decision, the steps themselves are often quickly forgotten. Within our framework, this information is specifically collected and stored in a database, so that users can trace back the decision-making process.


The experts' knowledge is fed into a semi-/supervised pipeline to calibrate the output of automated detection (\textbf{C4}). After developing a careful understanding of what the users needed, we based the refreshment process of the semi-/supervised pipeline on application-specific batch processing of annotations. We justify our design based on the high variability in anomaly frequency among domains. Based on the satellite company's particular needs, we concluded that a weekly update is sufficient.

\section{Evaluation}

In the following section, we introduce several experiments that demonstrate the use, performance, and effectiveness of \sintel. We experimentally evaluate its ability to complete anomaly detection in unsupervised and semi-supervised settings. In addition, we report the results of our study with the satellite company's operation team.  

\textbf{Datasets.}
Our experiments utilize three publicly-accessible time series datasets with known anomalies.
First, we use two sets of spacecraft telemetry signals --- MSL and SMAP --- provided by \textbf{NASA},\footnote{NASA: \url{https://github.com/khundman/telemanom}} which contain a total of 80 signals and over 100 known anomalies.
Second, we use \textbf{Yahoo S5},\footnote{YAHOO S5 \url{https://webscope.sandbox.yahoo.com/catalog.php?datatype=s&did=70}} a dataset dealing with production traffic in Yahoo computing systems.
\review{Most subsets within this dataset have been synthetically created.} 
Overall, this dataset contains 367 signals and 2,152 anomalies.
Finally, we use the Numenta Anomaly Benchmark (\textbf{NAB}) dataset,\footnote{NAB: \url{https://github.com/numenta/NAB}} which has 45 signals with 94 anomalies.
\review{Table~\ref{tab:data_summary} summarizes each of the datasets used.}


\textbf{Experimental Setup.}
We first run the benchmark end-to-end on all available pipelines, using the described datasets.
We use a total of 6 pipelines -- LSTM DT~\cite{Hundman2018}, ARIMA~\cite{pena2013anomalyarima}, TadGAN~\cite{geiger2020tadgan}, LSTM AE~\cite{Malhotra2016}, and Dense AE -- and a pipeline that uses the MS Azure anomaly detection service~\cite{microsoft2019time}.
We evaluate the performances of the pipelines according to the evaluation metric described in Section~\ref{sec:scoring}.
To run the benchmark, we use a private HPC cluster with 192GB of memory and two 24-core Intel Xeon CPUs.

\begin{table}[ht]
    \centering
    \vspace{0.1cm}
    \begin{tabular}{lccc}
    \toprule
  
    Dataset & \# Signals & \# Anomalies  & Avg. Signal Length\\
                                    \midrule
    \textbf{NAB}         &  45            & 94    & 6088   \\
    \textbf{NASA}        &  80            & 103   & 8686   \\
    \textbf{YAHOO}       &  367           & 2152  & 1561   \\
  \bottomrule
    \end{tabular}
    \caption{Dataset Summary: 492 signals and 2349 anomalies.}\label{tab:data_summary}
    \vspace{-1.1cm}
\end{table}

\newsavebox\CBox
\def\textBF#1{\sbox\CBox{#1}\resizebox{\wd\CBox}{\ht\CBox}{\textbf{#1}}}

\begin{table*}[!ht]
\centering
\resizebox{\linewidth}{!}{%
\begin{tabular}{lccccccccc}
\toprule                             
                                    & \multicolumn{3}{c}{NAB} & \multicolumn{3}{c}{NASA} & \multicolumn{3}{c}{YAHOO}\\
                                      \cmidrule(lr){2-4}\cmidrule(lr){5-7}\cmidrule(lr){8-10}
                                    &    F1 & precision & recall &    F1 & precision & recall &    F1 & precision & recall \\
\midrule
LSTM DT                             & 0.555 $\pm$ 0.12 &  0.452 $\pm$ 0.09 & 0.734 $\pm$ 0.22 
                                    & 0.559 $\pm$ 0.15 &  0.472 $\pm$ 0.18 & 0.700 $\pm$ 0.08 
                                    & \textBF{0.772 $\pm$ 0.14} &  \textBF{0.880 $\pm$ 0.14} & 0.716 $\pm$ 0.21 \\
Dense AE                            & 0.599 $\pm$ 0.12 &  \textBF{0.666 $\pm$ 0.14} & 0.547 $\pm$ 0.10
                                    & \textBF{0.636 $\pm$ 0.10} &  \textBF{0.707 $\pm$ 0.09} & 0.578 $\pm$ 0.11 
                                    & 0.431 $\pm$ 0.40 &  0.793 $\pm$ 0.19 & 0.383 $\pm$ 0.40 \\
LSTM AE                             & \textBF{0.640 $\pm$ 0.09} &  0.632 $\pm$ 0.08 & 0.654 $\pm$ 0.11 
                                    & 0.583 $\pm$ 0.11 &  0.568 $\pm$ 0.09 & 0.601 $\pm$ 0.14 
                                    & 0.524 $\pm$ 0.26 &  0.760 $\pm$ 0.14 & 0.469 $\pm$ 0.33 \\
TadGAN                              & 0.606 $\pm$ 0.10 &  0.536 $\pm$ 0.10 & 0.721 $\pm$ 0.15 
                                    & 0.556 $\pm$ 0.12 &  0.473 $\pm$ 0.10 & 0.673 $\pm$ 0.16
                                    & 0.568 $\pm$ 0.19 &  0.698 $\pm$ 0.12 & 0.507 $\pm$ 0.26 \\
ARIMA                               & 0.514 $\pm$ 0.12 &  0.476 $\pm$ 0.14 & 0.566 $\pm$ 0.11 
                                    & 0.381 $\pm$ 0.07 &  0.383 $\pm$ 0.10 & 0.380 $\pm$ 0.05 
                                    & 0.757 $\pm$ 0.05 &  0.852 $\pm$ 0.14 & 0.714 $\pm$ 0.15 \\
MS Azure                            & 0.149 $\pm$ 0.11 &  0.086 $\pm$ 0.07 & \textBF{0.892 $\pm$ 0.11}
                                    & 0.041 $\pm$ 0.02 &  0.021 $\pm$ 0.01 & \textBF{0.873 $\pm$ 0.09} 
                                    & 0.494 $\pm$ 0.21 &  0.352 $\pm$ 0.18 & \textBF{0.912 $\pm$ 0.10} \\
\bottomrule
\end{tabular}%
}
\caption{\review{Unsupervised} anomaly detection results (F1 score, precision, and recall) per pipeline on each dataset.}
\label{tab:benchmarking}
\vspace{-20pt}
\end{table*}

\begin{figure*}[!ht]
    \centering
    \includegraphics[width=\linewidth]{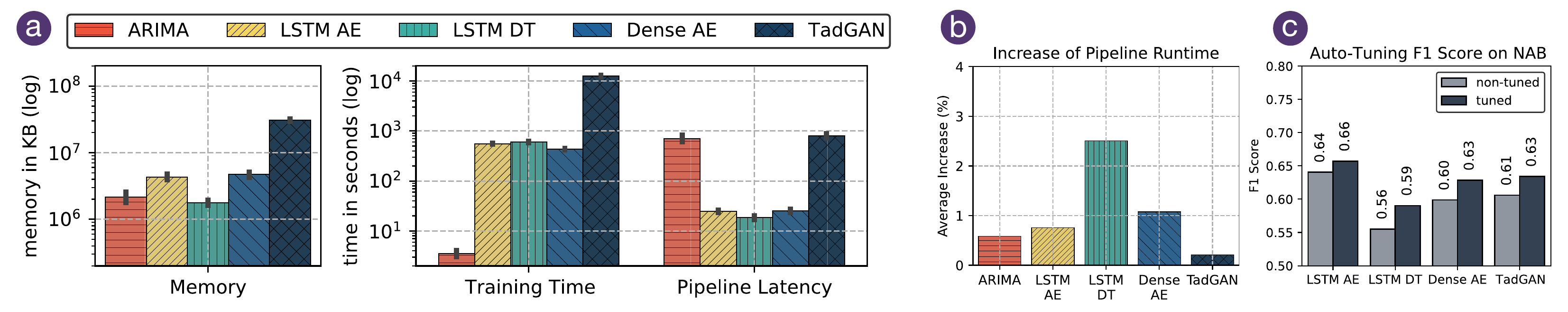}
    \vspace{-0.8cm}
    \caption{\review{(a) Pipeline computational performance.} (b) Difference in runtime between stand-alone primitives and end-to-end pipelines. \review{(c) F1 Scores prior to and after tuning pipelines on the NAB dataset using a ground truth set of anomalies.}}
    \vspace{-0.5cm}
    \label{fig:results_bigfig}
\end{figure*}

\textbf{Quality Performance.}
Table \ref{tab:benchmarking} shows the quality performances of currently available pipelines. The scores are calculated according to the \textit{overlapping segment} approach described in Section~\ref{sec:scoring}. 
As the table shows, no single pipeline outperforms all other pipelines -- each dataset has its own properties that make particular pipelines well- or ill-suited for it.
For example, MS Azure~\cite{microsoft2019time} manages to locate anomalies in all datasets, but at the expense of introducing many false positives. This number of false alarms could be prohibitively time-consuming for an expert monitoring \mbox{team to investigate.}

\textbf{Computational Performance.}
Figure~\ref{fig:results_bigfig}a shows the \textit{training time} \review{--- the time necessary to train the pipeline end-to-end}; the \textit{pipeline latency} \review{--- the time it takes the pipeline to produce an output while in \texttt{detect} mode}; and the \textit{memory} usage necessary for benchmarking all 462 signals for each of the pipelines presented.
We note that the TadGAN, LSTM AE, and Dense AE pipelines require the most memory due to their reconstructive natures.
\review{TadGAN takes the longest amount of time to train and produce outputs, likely due to its architecture: It is a GAN structure with four interleaved neural networks being trained simultaneously.}
ARIMA --- a popular statistical model --- requires a similar amount of time as deep learning pipelines once both training time and pipeline latency are factored in.
Users may be better served by different pipelines depending on their resources. Providing an assessment of the computational needs of each pipeline is necessary to help users determine which methods are the most appropriate for their particular case, especially when tackling deep learning models.  This important evaluative feature is missing from other current systems. 

\textbf{Primitive profiling.}
We evaluate the extra computational cost of using pipelines in our framework.
We first compute the total runtime required for each primitive to run in an external setting (outside of our framework). Next, we compare it to the time needed to run a pipeline from beginning to end. We determine the runtime of each model on the entire dataset (462 signals). We compute the delta as the difference between using a pipeline and running the primitives independently.
Although running primitives independently is faster than running the same primitives as part of a pipeline counterpart, the delta is generally minimal ($\mu \pm \sigma, \%$ avg. inc. time): ARIMA ($4.5\pm5.4s$, $0.58\%$), LSTM AE ($12.8\pm32.4s$, $0.75\%$), LSTM DT  ($15.6\pm17.6s$, $2.5\%$), Dense AE  ($17.8\pm44.4s$, $1.0\%$), and TadGAN ($28.7\pm46.4s$, $0.2\%$).
Figure~\ref{fig:results_bigfig}b illustrates the average percentage increase that comes from running primitives in our pipeline versus independently.
Given their stochastic nature, deep learning models tend to be more volatile from one signal to another, leading to a higher variability in runtimes.

\review{\textbf{AutoML performance.} \sintel improves pipelines using the hyperparameter optimization component introduced in Section 3.3. To test the efficacy of the platform's AutoML, we measure the F1 score per signal on the NAB dataset in a \textit{supervised} manner.
Pipelines improve $6.6\%$ on average. Figure~\ref{fig:results_bigfig}c shows the improvement in performance for each deep learning pipeline. Overall, $15\%$ of hyperparameter changes were in the postprocessing engine, specifically in the \texttt{find\_anomalies} primitive (see Figure~\ref{fig:orion-api}a). This demonstrates the level of
effectiveness of automated hyperparameter optimization that the user may expect to obtain.}

\textbf{Feedback evaluation.}
To validate the impact of the user feedback loop, we assess its performance at integrating user annotations by simulating human actions. Here we assume that a user has the capacity to annotate $k=2$ events at a single iteration and is capable of performing two types of annotations: adding or removing an event. The simulation stops when no events are left.

\begin{figure}[!b]
    \centering
    \vspace{-0.6cm}
    \includegraphics[width=\linewidth]{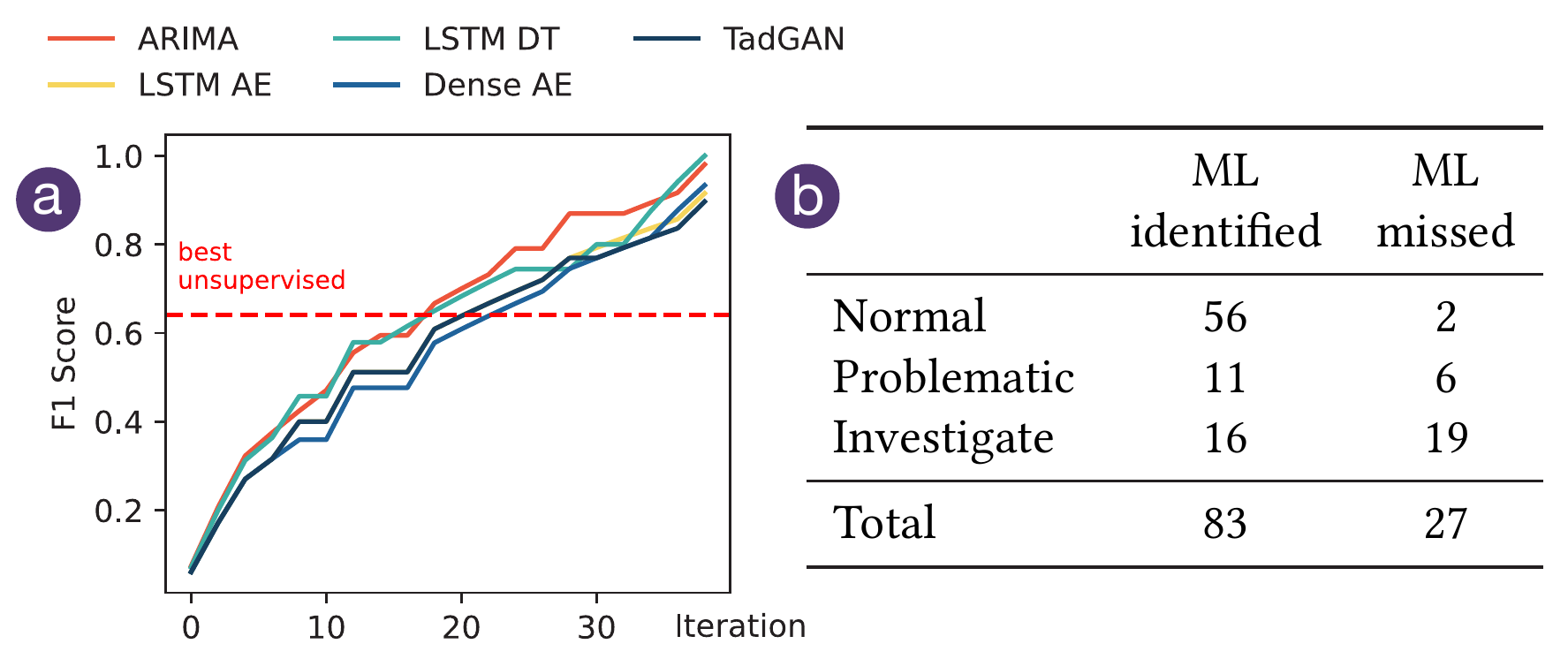}
    \vspace{-0.8cm}
    \caption{\review{(a) Semi-supervised pipeline performance on NAB through simulating annotations from different starting points. (b) Collected tags from real-world use case.}}
    \label{fig:tab:annotations}
    \vspace{-0.9cm}
\end{figure}

Our pipeline is a semi-supervised LSTM pipeline trained on sequences that were verified to be either anomalous or normal by the annotator. \review{We warm-start the simulation process with multiple initializations (all unsupervised pipelines)}. We use a 70/30 data split on the NAB dataset for training and testing. The training data encompasses 70 events, while the test data has 32 events. Results are depicted in Figure~\ref{fig:tab:annotations}a, where we observe that the performance of a semi-supervised pipeline surpasses \review{the best-performing} unsupervised pipeline once sufficient annotations have been obtained.

One drawback to depending solely on a semi-supervised pipeline is that before the pipeline becomes capable of identifying anomalies, its F1 Score is inferior to that of an unsupervised pipeline. Thus, we require a combination of unsupervised and semi-supervised pipelines to work synchronously. In addition, observing several flat segments in Figure~\ref{fig:tab:annotations}a, we note that some annotations may not help to improve detection. Given that all events must be annotated, it would be valuable to decide when retrain the pipeline by estimating the benefit gain ahead of time, so as not to incur unnecessary costs. 

\textbf{Real-world use case.}
We demonstrate the usability of our framework through the real-world use case introduced in Section~\ref{sec:example_scenario}.
We selected 16 real signals spanning a period of over 5 years from our collaborator's spacecraft telemetry database. These signals came from various subsystems and track metrics like electrical power, thermal temperature, etc.
We recorded the usage and activity of 6 experts and conducted interviews to gain their qualitative feedback.

We summarize the annotations collected from the study in Figure~\ref{fig:tab:annotations}b. A sample of 110 events tagged by humans were traced back a posteriori to determine whether or not the framework had also identified them as anomalous. The table depicts the events' detailed tags. The first column refers to the events identified by the ML model and then presented to domain experts for verification. The second column refers to the events that were missed by the ML model, but experts marked as worth detecting.
Among $110$ events total, the team deemed $52.7\%$ to be normal, confirmed 11 anomalies, manually added 6 events, and marked the rest as in need of further investigation. 
This illustrates the importance of incorporating human knowledge. Overall, experts valued our framework and believed in its ability to effectively identify and analyze anomalies.
\vspace{-0.5cm}
\section{Discussion}
In this section, we highlight some of the salient aspects of \sintel, its limitations, and our attempts to create a framework that could be improved and evolve over time.

\textbf{Why do we need humans in the loop?}
Unsupervised models are not perfect, particularly when past labels do not exist. Humans are necessary to iteratively guide even state-of-the-art models. In Figure~\ref{fig:tab:annotations}b, we note that the ML missed $\sfrac{27}{110}$ events. When investigating why, we discovered that some events, such as a lunar eclipse, have a normal shape, but should still be marked by experts for future reference. Meanwhile, some events, such as maneuvers, are actually considered normal by domain experts even though they have peculiar shapes. These issues are domain-specific, and it is difficult to find and understand them without a human annotator.

\textbf{Addressing distribution shifts.}
Any pipeline's performance relies on the \textit{preprocessing engine}. For example, domain experts pointed out that the ML detected events that were artifacts of aggregation.
Also, in our experiments, we witnessed a drop in F1 scores using unsupervised pipelines when detecting anomalies in Yahoo's A4 subset. Upon investigation, we discovered that 86\% of the signals in A4 contain a change point, which indicates a significant change in the data distribution. This could be overcome by preprocessing the signal using feature shift-elimination techniques such as decomposition~\cite{cleveland1990stldecomposition2, de2011forecastingdecomposition} as well as by segmenting signals using change point detection techniques~\cite{aminikhanghahi2017surveychangepoint, surveychangepoint2, burg2020evaluationchangepoint3}. Since \sintel{'s} pipelines are modular and composed of primitives, it is possible to add new preprocessing primitives and to integrate them into \sintel.

\review{\textbf{Mixing supervised and unsupervised.}}
\review{In Figure~\ref{fig:tab:annotations}a, we observed that the semi-supervised models initially performed worse than the unsupervised models. To mitigate this problem, we can couple the models together -- curating annotations as we collect them, so that we then have labeled data with which to train supervised models (pipelines). In subsequent iterations, we can run both supervised and unsupervised pipelines simultaneously as proposed in~\cite{veeramachaneni2016ai}.  As we saw in Figure~\ref{fig:tab:annotations}a, the model did not always benefit from new annotations, necessitating the need for such curation. We also note that pipelines will need to be updated when we observe drifts in the streaming data~\cite{webb2017understanding-distribdrift, wang2015conceptdrift}.}

\review{\textbf{Going beyond satellite operations.}}
\review{Although \sintel has been designed and implemented to address the needs of a satellite company operations team, it can be adapted for other applications with different data volumes and efficacy needs. In fact, we collaborated with one of the world's largest electric utility companies to inform our design. They have subsequently put \sintel to use, predicting component failures in wind turbines using application-specific pipelines. Most pipelines here are supervised (see Figure~\ref{fig:orion-api}b as an example) due to the availability of labels. You can find our wind energy repository available at \url{https://github.com/sintel-dev/Draco}.}

\review{\textbf{How will \sintel enable future research?}
\system provides a base to explore more avenues of research; for example, developing new methods of incorporating user feedback by leveraging supervised and unsupervised models together. It provides a benchmarking framework and a pipeline hub that will aid researchers in developing new ML models and comparing them to existing pipelines.
\sintel provides a way to collect and incorporate human feedback continuously, leading to new innovations made possible by human-in-the loop systems. For example, collecting human annotations can result in the design of new preprocessing primitives. }
\vspace{-0.8cm}
\section{Related Work}\label{sec:related_work}


\textbf{Time series AD algorithms.}
Many algorithms have been proposed to address time series anomaly detection~\cite{Chandola2009, Goldstein2016}. The most basic approaches simply flag regions where values exceed a certain threshold~\cite{Martinez-Heras2014, Decoste1997}.
More advanced methods are based on statistical hypothesis testing~\cite{zheng2016self}, clustering~\cite{Iverson2004, Iverson2008}, and/or machine learning~\cite{Yairi2006}.
Recent advancements in deep neural networks have led to the emergence of deep learning-based anomaly detection approaches~\cite{Malhotra2016, Hundman2018, geiger2020tadgan}. However, in order to be effectively used by domain experts, these algorithms cannot stand alone and must become part of an end-to-end workflow with relevant APIs.

\textbf{Time series AD systems.}
A wide range of systems made specifically for time series data have been proposed. These address a variety of tasks including classification~\cite{smile-system}, feature extraction~\cite{christ2018tsfresh}, and anomaly detection~\cite{veeramachaneni2016ai, Laptev2015, microsoft2019time, gao2020alibaba}.
Table~\ref{tab:related_work} summarizes the features present in some of the existing open source frameworks for anomaly detection. 
While these systems handle time series data, most of them only support a single anomaly detection algorithm.
Moreover, they fail to support a human-in-the-loop workflow.

\textbf{Benchmarks.}
Benchmarking frameworks are necessary to evaluate and compare different anomaly detection models~\cite{coleman2017dawnbench}.
\citet{Numenta15} introduced the first open-source benchmark library for time series anomaly detection on 58 signals from various sources.
However, it is difficult to integrate new algorithms into the library. 
\citet{jacob2021exathlon} introduced Exathlon --- a benchmark framework for anomaly detection and explanation discovery. This work addresses intricacies regarding the establishment of time series benchmarks, including evaluation metrics and performance monitoring.
Currently, benchmarking frameworks have limited pipelines, and are not easily extendable.

\textbf{Active incorporation of user feedback.} 
After automatically detecting anomalies, we need to continuously recalibrate the detection model to adapt to anomalies identified by human experts. 
\citet{pelleg2004active} propose an active learning approach that generates the detected anomalies and prompts the expert to classify them. $AI^2$~\cite{veeramachaneni2016ai} uses a similar strategy with an additional layer of combining supervised and unsupervised output. 
\citet{das2016aad} suggest an iterative process that involves training a supervised model, surfacing the most outlying points for expert review, and updating the model and feature weights accordingly. 
Although all of these methods are promising, they have been developed specifically for tabular data and require prior knowledge of data distribution. 
\vspace{-0.1cm}
\section{Conclusion}
In this paper, we have introduced a new end-to-end interactive anomaly detection framework for domain-specific time series. 
We first described the underaddressed problems that our system tries to solve. We then discussed its different components --- several state-of-the-art machine learning pipelines as well as a feedback integration mechanism --- and how they interact with each other.
We demonstrated how effective our framework can be for practical tasks, and presented a use case with a real-world application.
Overall, we have shown how Sintel can bridge the gap between domain experts and machine learning engineers, thus contributing heavily to the field of interactive machine learning.

\sintel is a larger ecosystem that can perform many tasks, including time series classification, regression, forecasting, and anomaly detection. All libraries are currently publicly available, with a large community providing feedback, improvements, and contributions.

\vspace{-0.1cm}
\section*{Acknowledgments}
We thank SES S.A. and Iberdrola/ScottishPower for their invaluable insights, feedback, and their financial support.
We thank Arash Akhgari for his assistance in editing the graphics.
Furthermore, the authors would like to acknowledge the contributions of Ihssan Tinawi, Manuel Alvarez Campo, Hector Dominguez, Alex Geiger, and Plamen Valentinov Kolev.
Sarah Alnegheimish is supported by a scholarship from King Abdulaziz City for Science and Technology.

\clearpage

\bibliographystyle{ACM-Reference-Format}
\bibliography{main.bib}

\end{document}